\documentclass{llncs}

\usepackage{amsmath}
\usepackage{amssymb}
\usepackage{pst-tree}
\usepackage{color}
\usepackage{algorithm}
\usepackage[noend]{algpseudocode}
\usepackage{hyperref}

\usepackage{llncsdoc}

\usepackage{tikz}
\usetikzlibrary{external}
\usepackage{pgfplots}

\usetikzlibrary{calc,trees,positioning,arrows,chains,shapes.geometric,
    decorations.pathreplacing,decorations.pathmorphing,shapes,
    matrix,shapes.symbols}
    
\tikzstyle{fleche}=[->,  >=stealth', thick]
\tikzstyle{bfleche}=[<-,  >=stealth', thick]
\tikzstyle{cercle}=[circle, draw]

\tikzset{
		text style/.style={
 		sloped,
    	text=black,
    	font=\footnotesize,
    	above
   	}
}

\usepackage{graphicx}
\usepackage{epstopdf}
\usepackage[T1]{fontenc}
\usepackage[utf8]{inputenc}

\usepackage{url}

\newcommand{\ie}{\emph{i.e.}~}
\newcommand{\cf}{\emph{cf.}~}
\newcommand{\eg}{\emph{e.g.}~}
\newcommand{\al}{\emph{al.}~}
\newcommand{\espace}{\vspace*{0.2cm}}

\newcommand*{\lmsbrace}{\lbrace\!\!\lbrace}
\newcommand*{\rmsbrace}{\rbrace\!\!\rbrace}

\mathchardef\mhyphen="2D

\begin{document}
\newtheorem{mydef}{Definition}
\newtheorem{myex}{Example}
\newtheorem{myproof}{Proof}

\title{Discriminant chronicles mining\thanks{This research is supported by the PEPS project funded by the French National Agency for Medicines and Health Products Safety (ANSM)}}
\subtitle{Application to care pathways analytics}

\author{Yann Dauxais\inst{1} \and Thomas Guyet\inst{2} \and David Gross-Amblard\inst{1} \and Andr\'e Happe\inst{3}}
\authorrunning{Yann Dauxais et al.}

\institute{Rennes University 1/IRISA-UMR6074,\\
\email{yann.dauxais@irisa.fr}
\and
AGROCAMPUS-OUEST/IRISA-UMR6074
\and
CHRU BREST/EA-7449 REPERES}

\maketitle

\begin{abstract}
Pharmaco-epidemiology (PE) is the study of uses and effects of drugs in well defined populations. 
As medico-administrative databases cover a large part of the population, they have become very interesting to carry PE studies. Such databases provide longitudinal care pathways in real condition containing timestamped care events, especially drug deliveries. Temporal pattern mining becomes a strategic choice to gain valuable insights about drug uses.
In this paper we propose $DCM$, a new discriminant temporal pattern mining algorithm.
It extracts chronicle patterns that occur more in a studied population than in a control population.
We present results on the identification of possible associations between hospitalizations for seizure and anti-epileptic drug switches in care pathway of epileptic patients.
\keywords{temporal pattern mining, knowledge discovery, pharmaco-epidemiology, medico-administrative databases}
\end{abstract}

\section{Introduction}
Healthcare analytics is the use of data analytics tools (visualization
, data abstraction, machine learning algorithms, etc.) applied on healthcare data. The general objective is to support clinicians to discover new insights from the data about health questions.

Care pathways are healthcare data with highly valuable information. 
A care pathway designates the sequences of interactions of a patient with the healthcare system (medical procedures, biology analysis, drugs deliveries, etc.). 
Care pathways analytics arises with the medico-administrative databases opening. 
Healthcare systems collect longitudinal data about patients to manage their reimbursements. Such huge databases, as the SNIIRAM \cite{Moulis2015411} in France, is readily available and has a better population coverage than \textit{ad hoc} cohorts. Compared to Electronic Medical Records (EMR), such database concerns cares in real life situation over a long period of several years.

Such databases are useful to answer various questions about care quality improvement (\eg care practice analysis), care costs cutting and prediction, or epidemiological studies. Among epidemiological studies, pharmaco-epidemiological (PE) studies answer questions about the uses of health products, drugs or medical devices, on a real population. 
Medico-administrative databases are of particular interest to relate care pathway events to specific outcomes. For instance, Polard et \al \cite{polard2015brand} used the SNIIRAM database to assess associations between brand-to-generic anti-epileptic drug substitutions and seizure-related hospitalizations. Dedicated data processing identified drug substitutions and statistical test assessed their relation to seizure-related hospitalization event. 

The main drawbacks of such an approach is that 1) epidemiologists often have to provide an hypothesis to assess, 2) they can not handle sequences with more than two events and 3) the temporal dimension of pathways are poorly exploited. Care pathway analytics will help to deeply explore such complex information source.
More especially temporal pattern mining algorithms can extract interesting sequences of cares, that would be potential study hypothesis to assess.

\espace

Temporal pattern mining is a research field that provides algorithms to extract interesting patterns from temporal data. These methods have been used in various medical data from time series to EMR datasets. 
Such method can be organized according to the temporal nature of the patterns they extract.
\textit{Sequential patterns} only takes into account the order of the events. It has been used in \cite{Wright201573} to identify temporal relationships between drugs. These relationships are exploited to predict which medication a prescriber is likely to choose next.
\textit{Temporal rules} \cite{concaro2009mining,berlingerio2007mining}, or more complex patterns like \textit{chronicles} \cite{huang2012mining,Alvarez2013}, model inter-event durations based on the event timestamps. 
Finally, \textit{Time interval patterns} \cite{MoskovitchKAIS15,Guyet2011} extract patterns with timestamps and durations.
For large datasets, the number of temporal patterns may be huge and not equally interesting. Extracting patterns related to a particular outcome enables to extract less but more significant patterns.
In \cite{lakshmanan2013investigating}, the temporal patterns are ranked according to their correlation with a particular patient outcome. 
Nonetheless, their number is not reduced. 

The frequency constraint is the limitation of these approaches. While dealing with PE study, pattern discriminancy seems more interesting to identify patterns that occur for some patients of interest but not in the control patients. Only few approaches proposed to mine discriminant temporal patterns \cite{Fradkin15,Quiniou2001}. Fradkin and M\"orchen \cite{Fradkin15} proposed the BIDE-D algorithm to mine discriminant sequential patterns and Quiniou et \al \cite{Quiniou2001} used inductive logic to extract patterns with quantified inter-event durations.

\espace

This article presents a temporal pattern mining algorithm that discovers discriminant chronicle patterns.
A chronicle is a set of events linked by quantitative temporal constraints.
In constraint satisfaction domain, chronicle is named temporal constraint network \cite{dechter1991temporal}. 
It is discriminant when it occurs more in the set of positive sequences than in the set of negative sequences.
To the best of our knowledge, this is the first approach that extracts discriminant pattern with quantitative temporal information. 
This algorithm is applied to pursue the Polard et \al analysis \cite{polard2015brand}. It aims at identifying possible associations between hospitalizations for seizure and anti-epileptic drug switch from SNIIRAM care pathways of epileptic patients.

\section{Discriminant chronicles} \label{sec:disciminant}
This section introduces basic definitions and defines the discriminant chronicle mining task.

\subsection{Sequences and chronicles} \label{sec:chronicles}
Let $\mathbb{E}$ be a set of event types and $\mathbb{T}$ be a temporal domain where $\mathbb{T} \subseteq \overline{\mathbb{R}}$. An \textbf{event} is a couple $(e,t)$ such that $e \in \mathbb{E}$ and $t \in \mathbb{T}$. We assume that $\mathbb{E}$ is totally ordered by $\leq_{\mathbb{E}}$.
A \textbf{sequence} is a tuple $\langle SID, \langle (e_1,t_1), (e_2,t_2), ..., (e_n,t_n) \rangle, L \rangle$ where $SID$ is the sequence index, $\langle (e_1,t_1), (e_2,t_2), ..., (e_n,t_n) \rangle$ a finite event sequence and $L\in \lbrace+,-\rbrace$ a label. 
Sequence items are ordered by $\prec$ defined as $\forall i, j \in [1,n],\; (e_i,t_i)\prec (e_j,t_j) \Leftrightarrow t_i<t_j \vee (t_i=t_j \wedge e_i<_\mathbb{E} e_j)$.		
\begin{myex}[sequence set, $\mathcal{S}$]\label{ex:dataset}
Table \ref{tab:dataset} represents a set of six sequences containing five event types ($A$, $B$, $C$, $D$ and $E$) and labeled with two different labels. 

\begin{table}[tb]
\footnotesize
	\begin{center}
		\tabcolsep = 2\tabcolsep
		\begin{tabular}{ccc}
			\hline
			\textbf{SID} & \textbf{Sequence} & \textbf{Label }\\
			\hline
			$1$ & $(A,1)$, $(B,3)$, $(A,4)$, $(C,5)$, $(C,6)$, $(D,7)$& $+$ \\
			$2$ & $(B,2)$, $(D,4)$, $(A,5)$, $(C,7)$ & $+$ \\
			$3$ & $(A,1)$, $(B,4)$, $(C,5)$, $(B,6)$, $(C,8)$,$(D,9)$& $+$ \\
			$4$ & $(B,4)$, $(A,6)$, $(E,8)$, $(C,9)$& $-$ \\
			$5$ & $(B,1)$, $(A,3)$, $(C,4)$& $-$ \\
			$6$ & $(C,4)$, $(B,5)$, $(A,6)$, $(C,7)$, $(D,10)$& $-$ \\
			\hline
		\end{tabular}
	\end{center}
	\caption{Set of six sequences labeled with two classes $\{+,-\}$.}
	\label{tab:dataset}
\end{table}
\end{myex}

A \textbf{temporal constraint} is a tuple $\left(e_1, e_2, t^-, t^+\right)$, also noted $e_1[t^-,t^+]e_2$, where $e_1, e_2 \in \mathbb{E}$, $e_1 \leq_{\mathbb{E}} e_2$ and $t^-, t^+ \in \mathbb{T}$, $t^- \leq t^+$. A temporal constraint $e_1[t^-,t^+]e_2$ is said satisfied by a couple of events $\left(\left(e,t\right), \left(e', t'\right)\right)$ iff $e = e_1$, $e' = e_2$ and $t'-t \in [t^-,t^+]$. 

A \textbf{chronicle} is a couple $(\mathcal{E}, \mathcal{T})$ where $\mathcal{E} = \lmsbrace e_1 ... e_n \rmsbrace$, $e_i \in \mathbb{E}$ and $\forall i,j, 1 \leq i < j \leq n$, $e_i \leq_{\mathbb{E}} e_j$, $\mathcal{T}$ is a temporal constraint set: $\mathcal{T}=\{e[a,b]e'\; | \; e, e' \in \mathcal{E},\; e\leq_\mathbb{E}e'\}$.
As the constraint $e[a,b]e'$ is equivalent to $e'[-b,-a]e$, we impose the order on items, $\leq_{\mathbb{E}}$, to decide which one is represented in the chronicle.  
The set $\mathcal{E}$ is a \textbf{multiset}, \ie $\mathcal{E}$ can contain several occurrences of a same event type.

\begin{figure}[tb]
	\centering
	\includegraphics[width=.7\textwidth,keepaspectratio]{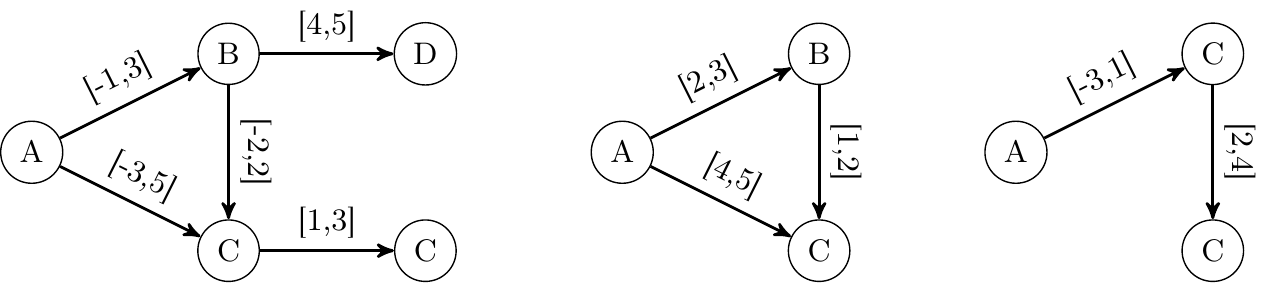}
	\caption{Example of three chronicles occurring in Table \ref{tab:dataset} (\cf Examples \ref{ex:chronique} and \ref{ex:support}). No edge between two events is equivalent to the temporal constraint $[-\infty,\infty]$.}
	\label{fig:chrondefexample}
\end{figure}

\begin{myex}\label{ex:chronique}
Fig. \ref{fig:chrondefexample} illustrates three chronicles represented by graphs.
Chronicle $\mathcal{C} = \left(\mathcal{E},\mathcal{T}\right)$ where 
$\mathcal{E}= \lmsbrace e_1 = A, e_2 = B, e_3 = C, e_4 = C, e_5 = D \rmsbrace$ and 
$\mathcal{T}=\{ e_1[-1,3]e_2,$ $e_1[-3,5]e_3,$ $e_2[-2,2]e_3, e_2[4,5]e_5,$ $e_3[1,3]e_4 \}$ is illustrated on the left. 
This graph is not complete. No edge between two events is equivalent to the temporal constraint $[-\infty,\infty]$, \ie there is no constraint.
\end{myex}

\subsection{Chronicle support}
Let $s = \langle (e_1,t_1), ..., (e_n,t_n) \rangle$ be a sequence and $\mathcal{C} = (\mathcal{E}=\lmsbrace e'_1, ..., e'_m \rmsbrace, \mathcal{T})$ be a chronicle.
An \textbf{occurrence} of $\mathcal{C}$ in $s$ is a sub-sequence $\tilde{s}= \langle (e_{f(1)},t_{f(1)}),$ $...,$ $(e_{f(m)},t_{f(m)}) \rangle$ such that 1) $f:\left[1,m\right]\mapsto[1,n]$ is an injective function, 2) $\forall i,\; e'_i=e_{f(i)}$ and 3) $\forall i,j,\; t_{f(j)}-t_{f(i)}\in [a,b]$ where $e'_i[a,b]e'_j \in \mathcal{T}$.
It is worth noting that $f$ is not necessarily increasing.
In fact, there is a difference between (i) the order of the chronicle multiset defined on items, $\leq_{\mathbb{E}}$, and (ii) the order on events in sequences, $\prec$, defined on the temporal domain.
The chronicle $\mathcal{C}$ \textbf{occurs} in $s$, denoted $\mathcal{C} \in s$, iff there is at least one occurrence of $\mathcal{C}$ in $s$.
The \textbf{support} of a chronicle $\mathcal{C}$ in a sequence set $\mathcal{S}$ is the number of sequences in which $\mathcal{C}$ occurs: $supp(\mathcal{C},\mathcal{S})=|\{s \in\mathcal{S}\;|\; \mathcal{C}\in s\}|$.
Given a minimal support threshold $\sigma_{min}$, a chronicle is \textbf{frequent} iff $supp(\mathcal{C},\mathcal{S}) \geq \sigma_{min}$.

\begin{myex}\label{ex:support}
Chronicle $\mathcal{C}$, Fig. \ref{fig:chrondefexample} on the left, occurs in sequences $1$, $3$ and $6$ of Table \ref{tab:dataset}.
We notice there are two occurrences of $\mathcal{C}$ in sequence $1$. 
Nonetheless, its support is $supp(\mathcal{C},\mathcal{S})=3$. This chronicle is frequent in $\mathcal{S}$ for any minimal support threshold $\sigma_{min}$ lower or equal to $3$.
The two other chronicles, so called $\mathcal{C}_1$ and $\mathcal{C}_2$ from left to right, occur respectively in sequences $1$ and $3$; and in sequence $6$. Their supports are $supp(\mathcal{C}_1, \mathcal{S})=2$ and $supp(\mathcal{C}_2, \mathcal{S})=1$.
\end{myex}

\subsection{Discriminant chronicles mining}
Let $\mathcal{S}^+$ and $\mathcal{S}^-$ be two sets of sequences, and $\sigma_{min}\in\mathbb{N}$, $g_{min}\in[1,\infty]$ be two user defined parameters. A chronicle is \textbf{discriminant} for $\mathcal{S}^+$ iff $supp(\mathcal{C},\mathcal{S}^+) \geq \sigma_{min}$ and $supp(\mathcal{C},\mathcal{S}^+) \geq g_{min}\times supp(\mathcal{C},\mathcal{S}^-)$. 
The \textbf{growth rate} $g(\mathcal{C},\mathcal{S})$ of a chronicle is defined by $\frac{supp(\mathcal{C},\mathcal{S}^+)}{supp(\mathcal{C},\mathcal{S}^-)}$ if $supp(\mathcal{C},\mathcal{S}^-) > 0$ and is $+\infty$ otherwise.
		
\begin{myex}
With chronicle $\mathcal{C}$ of Fig. \ref{fig:chrondefexample}, $supp(\mathcal{C},\mathcal{S}^+)=2$, $supp(\mathcal{C},\mathcal{S}^-)$ $=1$, where $\mathcal{S}^+$ (resp. $\mathcal{S}^-$) is the sequence set of Table \ref{tab:dataset} labeled with $+$ (resp. $-$). Considering that $g(\mathcal{C},\mathcal{S}) = 2$, $\mathcal{C}$ is discriminant if $g_{min} \leq 2$. 
For chronicles $\mathcal{C}_1$ and $\mathcal{C}_2$, $supp(\mathcal{C}_1,\mathcal{S}^+)=2$ and $supp(\mathcal{C}_1,\mathcal{S}^-)=0$  so $g(\mathcal{C}_1,\mathcal{S}) = +\infty$ and $supp(\mathcal{C}_2,\mathcal{S}^+)=0$ and $supp(\mathcal{C}_2,\mathcal{S}^-)=1$ so $g(\mathcal{C}_2,\mathcal{S}) = 0$. $\mathcal{C}_2$ is not discriminant, but $\mathcal{C}_1$ is for any $g_{min}$ value.
\end{myex}

The support constraint, using $\sigma_{min}$, prunes the unfrequent, and so insignificant, chronicles.
For example, a chronicle like $\mathcal{C}_1$ such that $g(\mathcal{C}_1,\mathcal{S}) = +\infty$ but  $supp(\mathcal{C}_1,$ $\mathcal{S}^-) = 0$ is discriminant but not interesting.
Pruning can be done efficiently thanks to the anti-monotonicity of frequency, which is also valid for chronicle patterns \cite{Alvarez2013}.  
More specifically, if a chronicle\footnote{$\mathcal{T}_{\infty}$ is the set of temporal constraints with all bounds set to $\infty$.} $(\mathcal{E}, \mathcal{T}_{\infty})$ is not frequent, then no chronicle of the form $\left(\mathcal{E},\mathcal{T}\right)$ will be frequent. This means that temporal constraints may be extracted only for frequent multisets.

\espace

Extracting the complete set of discriminant chronicles is not interesting. Discriminant chronicles with same multiset and similar temporal constraints are numerous and considered as redundant. 
It is preferable to extract chronicles whose temporal constraints are the most generalized.
The approach proposed in the next section efficiently extracts an incomplete set of discriminant chronicles that we want to be meaningful.
\section{$DCM$ algorithm} \label{sec:algo}

\begin{algorithm}[tb]
\caption{Algorithm $DCM$ for discriminant chronicles mining}
\label{algo:sketch}
\footnotesize
\begin{algorithmic}[1]
\Require $\mathcal{S}^+$, $\mathcal{S}^-$~: sequences sets, $\sigma_{min}$~: minimal support threshold, $g_{min}$~: minimal growth threshold
\State $ \mathbb{M} \gets $ \Call{ExtractMultiSet}{$\mathcal{S}^+$, $\sigma_{min}$}\Comment{$\mathbb{M}$ is the frequent multisets set}
\State{$\mathbb{C} \gets \emptyset$} \Comment{$\mathbb{C}$ is the discriminant chronicles set}

\ForAll{$ms \in \mathbb{M}$}
	\If{ $supp\left(\mathcal{S}^+, \left(ms,\mathcal{T}_{\infty}\right)\right) > g_{min}\times supp\left(\mathcal{S}^-, \left(ms,\mathcal{T}_{\infty}\right)\right)$ } \label{line:ms_discr}
		\State $ \mathbb{C} \gets \mathbb{C} \cup \{(ms,\mathcal{T}_{\infty})\}$\Comment{Discriminant chronicle without temporal constraints}
	\Else
		\ForAll{$\mathcal{T} \in $
		\Call{ExtractDTC}{$\mathcal{S}^+$, $\mathcal{S}^-$, $ms$, $g_{min}$, $\sigma_{min}$}}
		\State $\mathbb{C} \gets \mathbb{C} \cup \{(ms,\mathcal{T})\}$ \Comment{Add a new discriminant chronicle}
		\EndFor
	\EndIf
\EndFor
\State \Return $\mathbb{C}$
\end{algorithmic}
\end{algorithm}

The $DCM$ algorithm is given in Algorithm~\ref{algo:sketch}. 
It extracts discriminant chronicles in two steps: the extraction of frequent multisets, and then the specification of temporal constraints for each non-discriminant multiset.

At first, \Call{ExtractMultiSet}{} extracts $\mathbb{M}$, the frequent multisets in $\mathcal{S}^+$. It applies a regular frequent itemset mining algorithm on a dataset encoding multiple occurrences. An item $a \in \mathbb{E}$ occurring $n$ times in a sequence is encoded by $n$ items: $I^a_1, \dots, I^a_n$.
A frequent itemset of size $m$, $(I^{e_k}_{i_k})_{1\leq k \leq m}$, extracted from this dataset is transformed into the multiset containing, $i_k$ occurrences of the event $e_{k}$. 
Itemsets with two items $I_{i_k}^{e_k}$, $I_{i_l}^{e_l}$ such that $e_k=e_l$ and $i_k\neq i_l$ are redundant and thus ignored.

In a second step, lines 3 to 8 extract the discriminant temporal constraints (DTC) of each multiset.
The naive approach would be to extract DTC for all frequent multisets. A multiset  $\mathcal{E}$ (\ie a chronicle $\left(\mathcal{E},\mathcal{T}_\infty\right)$) which is discriminant may yield numerous similar discriminant chronicles with most specific temporal constraints. We consider them as useless and, as a consequence, line 4 tests whether the multiset $ms$ is discriminant. If so, $\left(ms,\mathcal{T}_\infty\right)$ is added to the discriminant patterns set. Otherwise, lines 7-8 generate chronicles from DTC identified by \Call{ExtractDTC}{}.

\subsection{Temporal constraints mining}
The general idea of \Call{ExtractDTC}{} is to see the extraction of DTC as a classical numerical rule learning task \cite{Cohen95}.
Let $\mathcal{E}=\lmsbrace e_1 ... e_n \rmsbrace$ be a frequent multiset. A relational dataset, denoted $\mathcal{D}$, is generated with all occurrences of $\mathcal{E}$ in $\mathcal{S}$. The numerical attributes of $\mathcal{D}$ are inter-event durations between each pair $(e_i,e_j)$, denoted $\mathcal{A}_{e_i\rightarrow e_j}$. Each occurrence yields one example, labeled by the sequence label ($L\in\{+,-\}$).

A rule learning algorithm induces numerical rules from $\mathcal{D}$. A rule has a label in conclusion and its premise is a conjunction of conditions on attribute values. Conditions are inequalities in the form: $\mathcal{A}_{e_i \rightarrow e_j} \geq x \wedge \mathcal{A}_{e_i \rightarrow e_j} \leq y $, where $(x, y) \in \overline{\mathbb{R}}^2$.
These rules are then translated as temporal constraints, $e_i[x, y]e_j$, that make the chronicle discriminant.

\begin{table}[tb]
\footnotesize
\centering
\begin{tabular}{ccccc}
\hline
SID & $\mathcal{A}_{A\rightarrow B}$ & \textbf{$\mathcal{A}_{B\rightarrow C}$} & \textbf{$\mathcal{A}_{A\rightarrow C}$} & Label\\
\hline 
\textbf{1} & $2$ & $2$& $4$& $+$ \\
\textbf{1} & $-1$ & $2$& $1$& $+$ \\
\textbf{2} & $5$ & $-2$& $3$& $+$ \\
\textbf{3} & $3$& $0$& $3$& $+$ \\
\textbf{5} & $-1$& $3$& $1$& $-$ \\
\textbf{6} & $6$ & $-1$& $5$& $-$ \\
\hline
\end{tabular}
\caption{Relational dataset for the multiset $\lmsbrace A,B,C\rmsbrace$.}
\label{tab:rulelearninginstance}
\end{table}

\begin{myex}
Table \ref{tab:rulelearninginstance} is the relational dataset obtained from the occurrences of $\lmsbrace A,B,C\rmsbrace$ in Table \ref{tab:dataset}.
The attribute $\mathcal{A}_{A\rightarrow B}$ denotes the durations between $A$ and $B$. We observe that several examples can come from the same sequence.

The rule $\mathcal{A}_{A\rightarrow B} \leq 5 \land \mathcal{A}_{B\rightarrow C} \leq 2 \implies +$ perfectly characterizes the examples labeled by $+$ in Table \ref{tab:rulelearninginstance}. It is translated by the DTC $\{A[-\infty, 5]B, B[-\infty,\allowbreak 2]C \}$ which gives the discriminant chronicle $\mathcal{C} = (\lmsbrace e_1 = A, e_2 = B, e_3 = C \rmsbrace, \allowbreak\left\{e_1 [-\infty, 5] e_2, e_2 [-\infty, 2] e_3 \right\})$.
\end{myex}

Our DCM implementation uses the Ripper algorithm \cite{Cohen95} to induce discriminant rules. Ripper generates discriminant rules using a growth rate computed on $\mathcal{D}$. 
To take into account multiple occurrences of a multiset in sequences, the growth rate of chronicles is reevaluated a posteriori on sequences datasets.
\section{Case study} \label{sec:casestudy}
This section presents the use of DCM to study care pathways of epileptic patients. Recent studies suggested that medication substitutions (so called switches) may be associated with epileptic seizures for patients with long term treatment with anti-epileptic (AE) medication. 
In \cite{polard2015brand}, the authors did not found significant statistical relationship between brand-to-generic substitution and seizure-related hospitalization.
The DCM algorithm is used to extract patterns of drugs deliveries that discriminate occurrences of recent seizure. These patterns may be interesting for further investigations by statistical analysis.

\subsection{Dataset}
Our dataset was obtained from the SNIIRAM database \cite{Moulis2015411} for epileptic patients with stable AE treatment with at least one seizure event. The treatment is said stable when the patient had at least 10 AE drugs deliveries within a year without any seizure. 
Epileptics seizure have been identified by hospitalization related to an epileptic event, coded G40.x or G41.x with ICD-10\footnote{ICD-10: International Classification of Diseases 10th Revision}.
The total number of such patients is 8,379. The care pathway of each patient is the collection of timestamped drugs deliveries from 2009 to 2011. 
For each drug delivery, the event id is a tuple $\langle m, grp, g\rangle$ where $m$ is the ATC code of the active molecule, $g\in\{0,1\}$ and $grp$ is the speciality group. The speciality group identifies the drug presentation (international non-proprietary name, strength per unit, number of units per pack and dosage form).
Our dataset contains 1,771,220 events of 2,072 different drugs and 20,686 seizure-related hospitalizations.

Since all patient sequences have at least one seizure event, we adapt the case-crossover protocol to apply our $DCM$ algorithm.
This protocol, consisting in using a patient as his/her own control, is often used in PE studies.
The dataset is made of two sets of 8,379 labeled sequences.
A $3$-days induction period is defined before the first seizure of each patient.
Drugs delivered within the $90$ days before inductions yield the positive sequences and those delivered within the $90$ days before the positive sequence, \ie the $90$ to $180$ days before induction, yield the negative sequences.
The dataset contains 127,191 events of 1,716 different drugs.

\subsection{Results}
Set up with $\sigma_{min} = 5.10^{-3}$, \ie $42$ patients\footnote{This number of patients have been initially estimated important by epidemiologists to define a population of patients with similare care sequences associated to seizure.}, and $g_{min} = 1.4$, we generated 777 discriminant chronicles.
Chronicles involved 510 different multisets and 128 different event types.

Three types of pattern are of specific interest for clinicians: (1) sequences of AE generic and brand-named drug deliveries, (2) sequences of same AE drug deliveries, (3) sequences with AE drug deliveries and other drug types deliveries.
According to these criteria, we selected $55$ discriminant chronicles involving $16$ different multisets to be discussed with clinicians.
For the sake of conciseness, we choose to focus the remaining of this section on chronicles related to \textit{valproic acid} (\textit{N03AG01} ATC code, with different presentations) but our results contain chronicles related to other AE drugs like \textit{levetiracetam} or \textit{lamotrigin}.

\subsubsection{Taking into account time in brand-to-generic substitution}

\begin{figure}[tb]
\centering
$\mathcal{C}_1$ \includegraphics[scale=0.29]{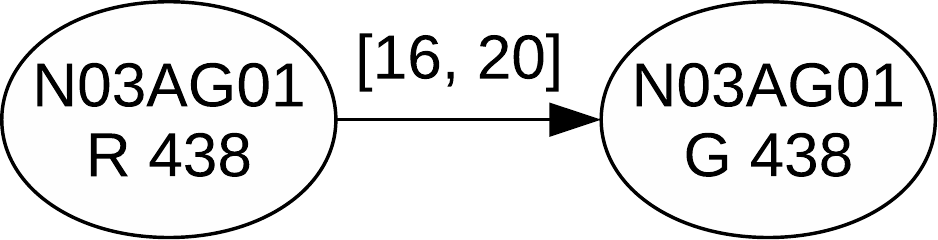} \hspace{0.2cm}
$\mathcal{C}_2$ \includegraphics[scale=0.29]{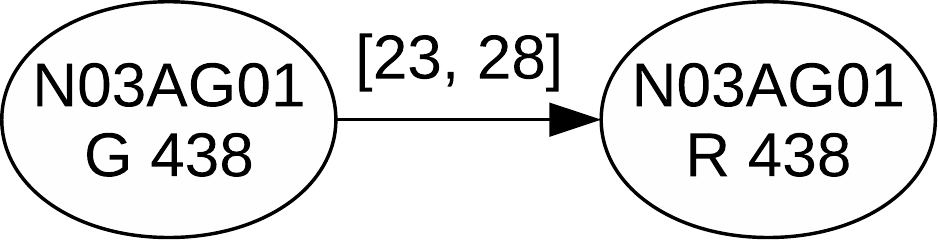}

\vspace*{0.1cm}

$\mathcal{C}_3$ \includegraphics[scale=0.29]{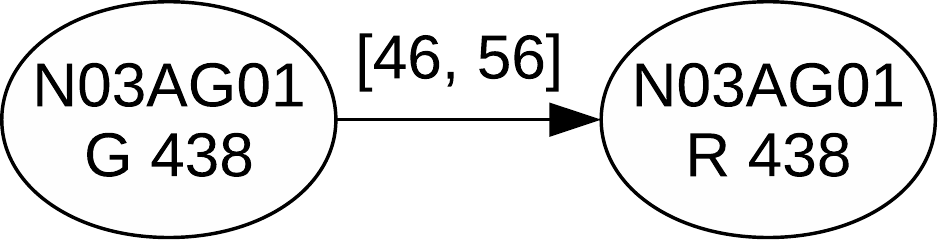} \hspace{0.2cm}
$\mathcal{C}_4$ \includegraphics[scale=0.29]{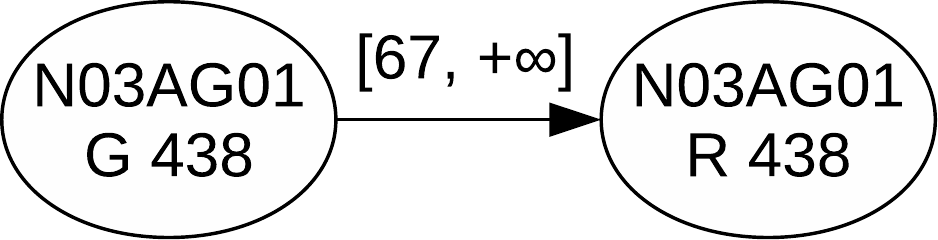}
\caption{Four discriminant chronicles describing switches between same type of valproic acid (N03AG01) generic (G 438) and brand-named (R 438).
$supp(\mathcal{C}_i,\mathcal{S}^+)$ respectively for $i=1$ to $4$ equals $43$, $78$, $71$ and $43$ and $supp(\mathcal{C}_i,\mathcal{S}^-)$ equals $23$, $53$, $39$ and $30$.}
\label{fig:gen_switch}
\end{figure}

We start with patterns representing switches between different presentation of \textit{N03AG01}. Fig.~\ref{fig:gen_switch} illustrates all discriminant patterns that have been extracted. 
It is noteworthy that all chronicles have temporal constraints, this means that multisets without temporal constraints are not discriminant.
This results is consistent with Polard et \al \cite{polard2015brand} which concluded that brand-to-generic AE drug substitution was not associated with an elevated risk of seizure-related hospitalization. But temporal constraints was not taken into account in the later.
The four extracted chronicles suggest that for some small patient groups, drug switches with specific temporal constraints are more likely associated to seizure.
 
The two first chronicles represent delivery intervals lower than $30$ days, respectively from brand-to-generic and generic-to-brand.
The third one represents an interval between the two events greater than $30$ days but lower than $60$ days.
The DTC of the last one could be interpreted as $[67,90]$ because of the bounded duration of the study period ($90$ days).
This chronicle represents a switch occurring more than $60$ days but most of the time less than $90$ days.

These behaviors may correspond to unstable treatments. In fact, AE drug deliveries have to be renew every months, thus, a regular treatment corresponds to a delay of $\approx 30$ days between two AE drug deliveries.

\espace

\begin{figure}[tb]
	\centering
	\includegraphics[scale=0.29]{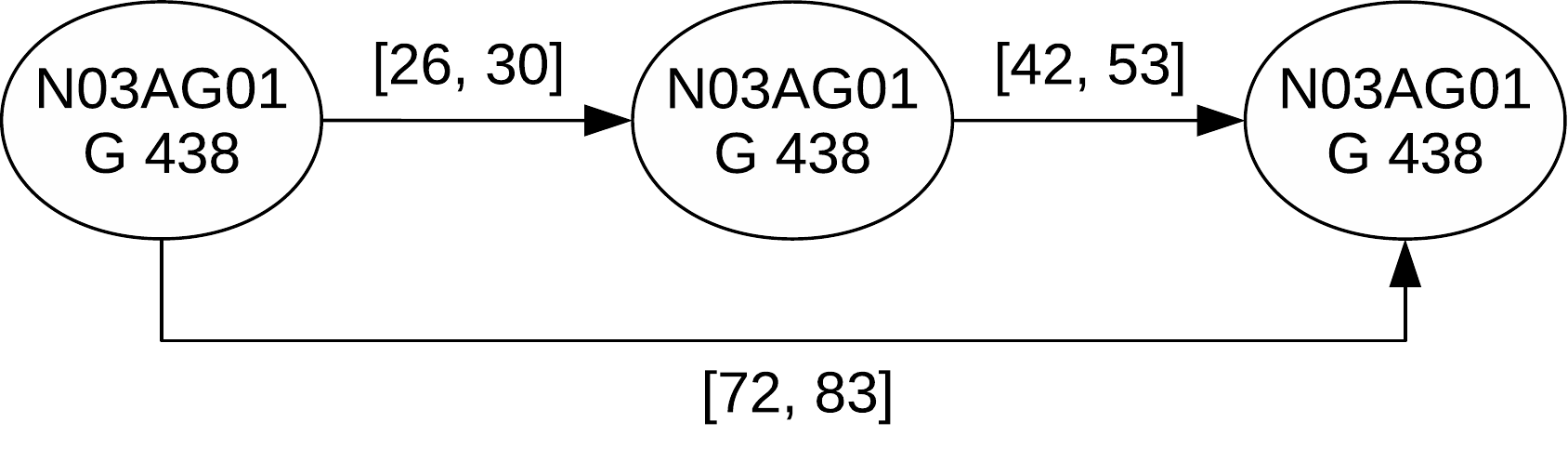}
	\vspace*{0.2cm}
	
	\includegraphics[scale=0.39]{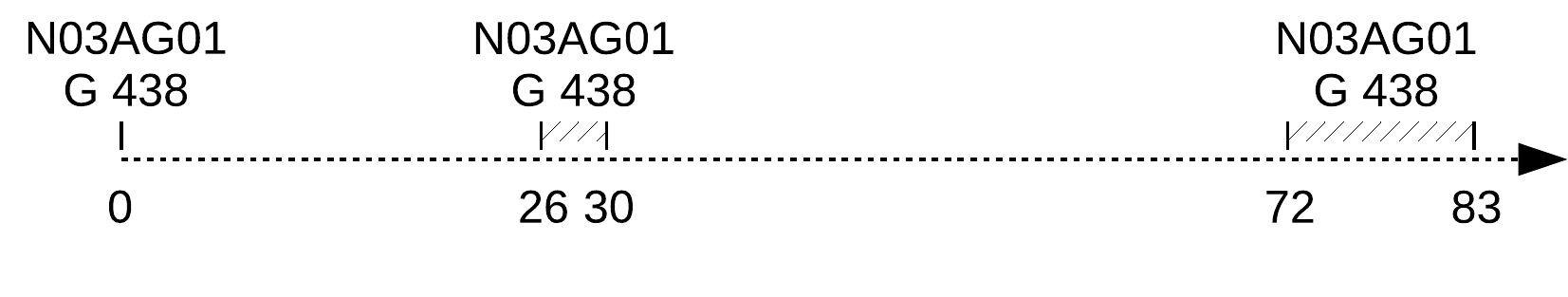}
	\caption{Above, a chronicle describing repetitions of \textit{valproic acid} (N03AG01) generic (G 438) and, below, its timeline representation. The chronicle is more likely related to epileptic seizure: $supp(\mathcal{C}, \mathcal{S}^+)=50$, $supp(\mathcal{C}, \mathcal{S}^-)=17$.}
	\label{fig:same_type}
\end{figure}

We next present in Fig. \ref{fig:same_type} an example of discriminant chronicle that involves three deliveries of \textit{N03AG01} (no chronicle involves more deliveries of this AE drug).

The growth rate of this chronicle is high ($2.94$). It is moreover simple to understand and, with their DTC, it can be represented on a timeline (see Fig. \ref{fig:same_type}, below). It is noteworthy that the timeline representation loses some constraint information. The first delivery is used as starting point ($t_0$), but it clearly illustrates that last delivery occurs too late after the second one (more $30$ days after).
As well as previous patterns, this chronicle describes an irregularity in deliveries. More precisely, the irregularity occurs between the second and the third deliveries as described by the DTC $[42,53]$ and $[72, 83]$.

We conclude from observations about the previous two types of patterns that the precise numerical temporal information discovered by $DCM$ is useful to identify discriminant behaviors.
Analyzing pure sequential patterns does not provide enough expression power to associate switch of same AE deliveries with seizure. 
Chronicles, specifying temporal constraints, allow us to describe the conditions under which a switch of same AE deliveries is discriminant for epileptic seizure.

\subsubsection{Example of a complex chronicle}

\begin{figure}[tb]
	\centering
	\includegraphics[scale=0.29]{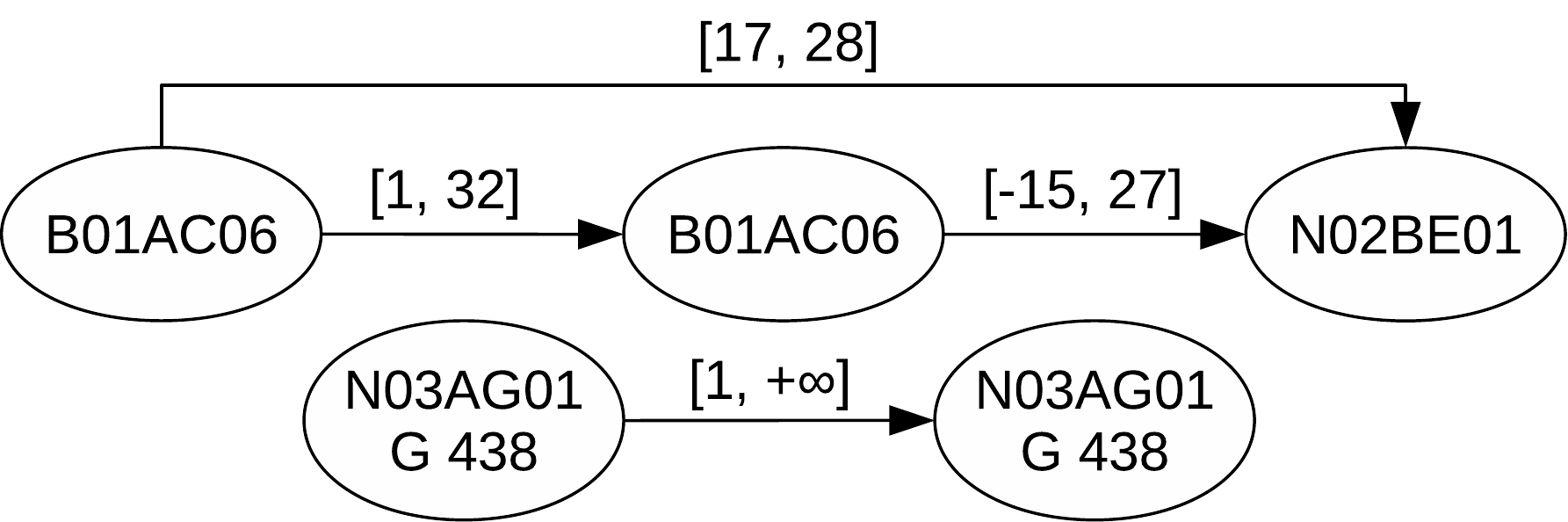}
	\caption{A chronicle describing co-occurrences between anti-thrombosis drugs (\textit{B01AC06}) and \textit{valproic acid} which is more likely associated to seizure: $supp(\mathcal{C}, \mathcal{S}^+)=42$, $supp(\mathcal{C}, \mathcal{S}^-)=20$.}
	\label{fig:other_epi}
\end{figure}

The chronicle presented in Fig. \ref{fig:other_epi} has been judged interesting by clinicians as a potential adverse drug interaction between an AE drug and a drug non-directly related to epilepsy, more especially aspirin (\textit{B01AC06}), prescribed as an anti-thrombotic treatment.
The $DTC$ implies that aspirin and paracetamol (\emph{N02BE01}) are delivered within a short period (less 28 days). There is no temporal relations between these deliveries and the deliveries of \textit{valproic acid}. But their co-occurrence within the period of $90$ days is part of the discriminatory factor.

After a deeper analysis of patient care pathways supporting this chronicle, clinicians made the hypothesis that these patients were treated for brain stroke. It is known to seriously exacerbate epilepsy and to increase seizure risk.

\section{Conclusion}
This article presents a new temporal pattern mining task, the extraction of discriminant chronicles, and the $DCM$ algorithm which extracts them from temporal sequences.
This new mining task appears to be useful to analyze care pathways in the context of pharmaco-epidemiology studies and we pursue the Polard et \al \cite{polard2015brand} study on epileptic patients with this new approach.
On the one hand, extracted patterns are discriminant. They correspond to care sequences that are more likely associated to a given outcome, \ie epileptic seizures in our case study. Such patterns are less numerous and more interesting for clinicians. 
On the other hand, the main contribution of $DCM$ algorithm is the discovery of temporal information that discriminates care sequences. Even if a sequence of care events is not discriminant (\eg drug switches), the way they temporally occur may witness the outcome (\eg seizure).

Experimental results on our case study show that $DCM$ extracts a reduced number of patterns. Discriminant patterns have been presented to clinicians who conclude of their potential interestingness by exploring the care pathways of sequences supported by chronicles. 
At this stage of the work, our main perspective is to integrate $DCM$ in a care pathways analytics tool such that extracted chronicles may easily be contextualized in care pathways and manually modified to progressively build an care sequence of interest.

\bibliographystyle{splncs03}
\bibliography{aime2017}

\end{document}